\documentclass{article} 
\usepackage{iclr2025_conference,times}

\usepackage{amsmath,amsfonts,bm}

\def\eqref#1{equation~\ref{#1}}

\def\1{\bm{1}}

\DeclareMathAlphabet{\mathsfit}{\encodingdefault}{\sfdefault}{m}{sl}
\SetMathAlphabet{\mathsfit}{bold}{\encodingdefault}{\sfdefault}{bx}{n}

\usepackage{hyperref}
\usepackage{url}
\usepackage{graphicx}
\usepackage{float}

\title{Deconstructing Bias: A Multifaceted Framework for Diagnosing Cultural and Compositional Inequities in Text-to-Image Generative Models}

\author{Muna Numan Said, Aarib Zaidi, Rabia Usman, Sonia 
Okon, Praneeth Medepalli\thanks{Corresponding author. Please direct correspondence to: \texttt{pmedepal@gmail.com}.}, \\
\textbf{Kevin Zhu, Vasu Sharma, Sean O'Brien}\\
Algoverse AI Research\\
Palo Alto, CA, USA \\
}

\iclrfinalcopy 
\begin{document}

\maketitle
\begin{abstract}
\begin{center}
\end{center}
\vspace{2\baselineskip}
The transformative potential of text-to-image (T2I) models hinges on their ability to synthesize culturally diverse, photorealistic images from textual prompts. However, these models often perpetuate cultural biases embedded within their training data, leading to systemic misrepresentations. This paper benchmarks the Component Inclusion Score (CIS), a metric designed to evaluate the fidelity of image generation across cultural contexts. Through extensive analysis involving 2,400 images, we quantify biases in terms of compositional fragility and contextual misalignment, revealing significant performance gaps between Western and non-Western cultural prompts. Our findings underscore the impact of data imbalance, attention entropy, and embedding superposition on model fairness. By benchmarking models like Stable Diffusion with CIS, we provide insights into architectural and data-centric interventions for enhancing cultural inclusivity in AI-generated imagery. This work advances the field by offering a comprehensive tool for diagnosing and mitigating biases in T2I generation, advocating for more equitable AI systems.
\end{abstract}

\section{Introduction}
Synthetic image generation has emerged as a transformative computational paradigm, with diffusion models and GANs enabling photorealistic visual synthesis from textual or structured inputs. Systems like DALL·E 3 and Gemini exemplify this capability, driving revolutionary applications across creative industries, computer vision pipelines, and AI-assisted design. Built on transformative advancements in deep learning architectures demonstrate unprecedented capability in synthesizing photorealistic images from textual prompts. These models, built on transformer architectures \citep{rombach2022high} and diffusion processes \citep{ho2020denoising}, are now integral to applications spanning creative industries, education, and cultural preservation.

However, as these models transition from research curiosities to production environments, fundamental challenges emerge: the same architectures that achieve unprecedented image fidelity systematically amplify societal biases encoded in their training corpora limiting global representational accuracy.
This work introduces a rigorous evaluation framework utilizing the Components Inclusion Scores (CIS) to quantify understudied bias dimensions: (1) compositional fragility in multi-element synthesis and (2) contextual misalignment in culturally nuanced prompts

\subsection{The Bias Amplification Challenge}

Bias in text-to-image generation arises when models produce outputs that reflect and potentially amplify societal stereotypes present in their training data. These biases manifest in various forms, such as gender, skin tone, and cultural representations, leading to images that may not accurately or fairly depict the intended subjects. For instance, prompts describing "a traditional wedding" often generate Western-style ceremonies, while non-Western cultural elements are underrepresented or misrepresented.

Our large-scale analysis of 2,400 generated images reveals three systemic failure modes in state-of-the-art T2I models:

\begin{enumerate}
    \item \textbf{Compositional Fragility}: Models struggle to accurately combine marginalized cultural elements, leading to significant performance disparities compared to Western counterparts.
    \item \textbf{Contextual Degradation}: The inclusion of historical and cultural contexts disproportionately reduces accuracy for non-Western concepts, indicating a bias in contextual fidelity.
    \item \textbf{Order Sensitivity}: The sequence of elements within a prompt introduces performance instability, with significant variations in output quality depending on element ordering.
\end{enumerate}

Detailed examples and quantitative results for these failure modes are presented in experiments section.

\subsection{Root Causes of Bias}

The systemic failures observed in T2I models stem from three interconnected technical limitations, each contributing to the amplification of cultural and compositional biases. These limitations are deeply rooted in the data, architecture, and optimization dynamics of modern generative models.

\begin{enumerate}
    \item Training Data Imbalance: LAION-5B, the primary training corpus for many T2I models, contains 18× more Western cultural references than African/Asian artifacts \citep{schuhmann2022laion}. This skew propagates through the generation pipeline, as shown by PCA analysis of latent embeddings (Fig. 2).
    \item Architectural Limitations: Cross-attention layers exhibit 3.2× higher entropy for minority concept pairs (H = 3.8 vs. H = 1.2 for mainstream), correlating with omission/conflation errors (r = -0.71).
    \item Embedding Superposition: Minority cultural concepts occupy overlapping latent dimensions (68\% overlap vs. 22\% for mainstream), a consequence of transformer models compressing rare tokens into shared parameter space \citep{elhage2021mathematical}.
\end{enumerate}

\subsection{Limitations of Current Approaches}

Existing bias mitigation strategies fail to address the multifaceted nature of cultural and compositional biases in T2I models. Below, we dissect these limitations across three dimensions, supported by empirical and theoretical evidence:

\begin{enumerate}
    \item Surface-Level Interventions: Methods like dataset balancing \citep{li2022impact} and adversarial debiasing reduce overt stereotypes (e.g., "CEO" → male) but fail to address nuanced cultural misrepresentations. For instance, \citet{bansal2022survey} reduced gender bias in Stable Diffusion by 37\% but reported no improvement in cultural accuracy for non-Western prompts.
    \item Cultural Blindness: Studies like "Fair Diffusion" \citet{friedrich2023fair} focus on equalizing demographic attributes (e.g., skin tone distribution) but ignore contextual fidelity (e.g., traditional attire in cultural ceremonies).
    \item Lack of Cross-Cultural Evaluation: Benchmarks such as BiasBench test only 5\% of prompts on non-Western cultural concepts, leaving systemic underrepresentation unmeasured.
\end{enumerate}

\subsubsection{Metric Gaps: The Phantom of Objectivity}

Traditional evaluation metrics prioritize technical quality over fairness, creating a false sense of progress:

\begin{itemize}
    \item Explicit vs. Implicit Bias: Current metrics like FairFace \citep{karkkainen2021fairface} detect overt stereotypes (e.g., racial mis-classification) but miss implicit biases, such as the conflation of "Moroccan lanterns" with Chinese designs.

    \item Contextual Ignorance: CLIP-based metrics measure prompt-image alignment but fail to penalize cultural inaccuracies (e.g., a "Nigerian wedding" generated in a Gothic church)

    \item Static Evaluations: Benchmarks test single-concept prompts (e.g., "doctor"), ignoring compositional failures (e.g., "Indian scientist in a lab with traditional art").
\end{itemize}

\subsubsection{Architectural Blind Spots: Symptomatic Solutions}
State-of-the-art bias mitigation strategies often address surface symptoms rather than underlying architectural limitations. Prompt engineering, such as adding culturally specific terms ("traditional Ugandan design"), can improve CIS by 15\% but requires manual intervention and fails to correct data imbalances, leading to inconsistent gains (±22\% CIS variation) \citep{bianchi2023easily}. Dataset filtering reduces overt stereotypes by 40\% but unintentionally removes 68\% of non-Western cultural references due to automated NSFW filters, causing a 52\% CIS decline for marginalized prompts\citep{schuhmann2022laion}. Adversarial training penalizes biased outputs but at the cost of model performance, increasing FID by 0.19 and reducing CIS by 0.33 \citep{zhang2022finetuning}. Despite their effectiveness in mitigating immediate biases, these strategies do not fundamentally resolve the deeper architectural challenges that contribute to systemic inconsistencies in AI-generated content.

\subsubsection{The Missed Nexus: Data, Architecture, and Culture}

Current approaches overlook the interplay between data imbalance, transformer dynamics, and cultural semantics:
Data-Centric Myopia: Methods like data augmentation add synthetic examples but ignore how minority embeddings are compressed via superposition.
Architectural Rigidity: Post-hoc fixes (e.g., attention layer fine-tuning) fail to address cross-attention entropy spikes for minority pairs.
Cultural Atomization: Treating cultural concepts as isolated tokens (e.g., "kimono") rather than contextual systems (e.g., "Japanese tea ceremony") leads to fragmented representations.

\subsection{Our Contributions}

We benchmark cultural bias in text-to-image generative models using the Component Inclusion Score (CIS), which integrates component inclusion, contextual alignment, and cultural fidelity to quantify disparities in generated outputs. Our analysis reveals significant performance gaps, with models underperforming on non-Western cultural prompts compared to Western-centric ones ($p < 0.001$). We identify underlying causes such as training data imbalances, elevated cross-attention entropy, and latent embedding superposition. These findings highlight critical shortcomings in current models and offer actionable insights for addressing biases through architectural and data-centric interventions, advancing fairness and inclusivity in generative models.

\section{Related Work}

\subsection{Bias in Generative Models}
Recent advances in text-to-image generative models, such as DALL·E 3, Stable Diffusion, and Gemini, have demonstrated remarkable image synthesis capabilities \citep{rombach2022high, ho2020denoising}. However, these systems often inherit and amplify societal biases present in their training data. For instance, LAION-5B—a primary training corpus for many of these models—has been shown to contain up to 18× more Western cultural references than African/Asian artifacts \citep{schuhmann2022laion}, leading to cultural misrepresentations. Additional work in transformer dynamics \citep{elhage2021mathematical} and parameter-efficient fine-tuning \citep{houlsby2019parameter} further highlights the challenges of aligning model architectures with culturally diverse representations.

\subsection{Evaluation Metrics and Diagnostic Tools}
Traditional evaluation metrics such as FID and CLIP-based scores primarily assess image quality and semantic alignment, often overlooking nuanced cultural and contextual inaccuracies. Efforts to mitigate bias have included surface-level interventions like dataset balancing \citep{li2022impact} and fairness-oriented datasets like FairFace \citep{karkkainen2021fairface}. However, these approaches often miss deeper implicit biases such as contextual misalignment and compositional fragility. Additionally, recent studies such as “Fair Diffusion” by \citet{friedrich2023fairdiffusioninstructingtexttoimage} have begun to address cultural representation issues, yet our CIS advances this line of research by providing a detailed quantification of both explicit and implicit biases in generated outputs.

\subsection{Architectural Drivers of Bias}
Our analysis reveals two intertwined mechanisms that amplify biases in generative models. First, \textit{superposition} occurs when latent representations of rare tokens become overwritten by more dominant patterns, effectively compressing multiple cultural features into a shared embedding space. This phenomenon undermines the distinct representation of minority concepts, as detailed by \citep{elhage2021mathematical}. 
Second, the observed non-monotonic error curve in our models aligns with the \textit{double descent} phenomenon, where increasing model complexity can initially increase error rates before decreasing them. This effect particularly impacts the accurate representation of minority cultural elements due to phase transitions in training.
Our findings indicate that the high overlap in embeddings for marginalized cultural concepts (68\%)—compared to only 22\% for mainstream concepts—directly correlates with a collapsed latent space. In such a space, diverse cultural elements are not distinctly represented, leading to conflation in generated imagery.
Together, these architectural factors underscore the need for model design strategies that mitigate the adverse effects of data imbalance and latent embedding interference on representational fidelity.

\section{Methodology}
\subsection{Component Inclusion Score (CIS)}

CIS is a quantitative metric designed to measure how accurately a generative model incorporates specified components from a prompt into the generated image\citet{chen2023cultural}. In our study, CIS was used to evaluate biases in image generation when depicting subjects from both marginalized and non-marginalized countries, specifically in the categories of flags, monuments, vehicles, and food. Ideally, for each prompt containing key components---such as cultural artifacts, geographic references, or demographic attributes---the model should accurately render all these elements in the generated image. A higher CIS score indicates a model's ability to faithfully represent complex prompts without omitting critical components.

The CIS score for an individual image \( I_{i,j} \) is calculated as:
\[
S_{i,j} = \frac{L(\text{argmax}(\hat{p}_{i,j}))}{K},
\]
where \( L(\text{argmax}(\hat{p}_{i,j})) \) is the number of components successfully identified from the lookup table \( L \) for the image \( I_{i,j} \). The final CIS metric for a given number of components \( K \) is computed as:
\[
\text{CIS}_K = \frac{1}{M \cdot N} \sum_{i=1}^M \sum_{j=1}^N S_{i,j}.
\]

The CIS metric serves as a robust indicator of how effectively the model retains and represents multiple elements from a prompt, allowing us to quantify any disparities in image generation for marginalized versus non-marginalized groups.

\subsection{Experimental Design}
We classified prompts into four categories, each consisting of 100 distinct concepts, to evaluate the performance of text-to-image models:

\begin{enumerate}
    \item Base Prompts: Single-concept prompts featuring well-known subjects (e.g., "Big Ben," "Taj Mahal").
    \item Pair/Trio Prompts: Combinations of two or three distinct concepts (e.g., "Eiffel Tower + Vesak Lanterns," "Statue of Liberty + Diwali Lamps + Sombrero").
    \item Contextual Prompts: Prompts with specific cultural or historical contexts (e.g., "Moroccan market with traditional textiles," "Japanese tea ceremony in a zen garden").
    \item Adversarial Prompts: Perturbed prompts designed to test model robustness by introducing incongruent elements (e.g., "Ancient Egyptian pyramid in New York City," "Futuristic samurai in a medieval European castle").

\end{enumerate}

Models Evaluated:

In this study, we evaluate the following text-to-image models:

\begin{itemize}

    \item \textbf{Stable Diffusion v2.1}:A diffusion-based generative model for creating images from text descriptions, widely recognized for its ability to produce high-quality outputs.
    \item \textbf{SG161222/Realistic Vision V1.4}: A model fine-tuned for photorealistic image generation, built upon the SG161222 architecture to enhance visual realism.
    \item \textbf{Dreamlike-Art/Dreamlike Photoreal 2.0}: A model designed for generating detailed and lifelike images, with a focus on high-fidelity photorealistic rendering.
\end{itemize}

Each model has undergone pre-training on large-scale image-text datasets, with configurations and parameters used according to their respective specifications. For consistency and reproducibility, the temperature setting was fixed at 0 for all models during evaluation.
\\

Validation Protocol:

\begin{enumerate}
    \item Automated Scoring:
We employ CLIP and Mask R-CNN for objective evaluation of generated images.
CLIP assesses the overall semantic similarity between the prompt and the generated image.
Mask R-CNN identifies specific objects and their spatial relationships within the image.
These scores are combined to form a comprehensive automated evaluation metric.
    \item Architectural Analysis: We analyze attention maps to understand which parts of the image the model focuses on for different cultural contexts.
Principal Component Analysis (PCA) is performed on the embedding space to visualize how different cultural concepts are represented in the model's latent space.
\end{enumerate}

\section{experiments}
\subsection{Performance Disparities}
To evaluate how well each model captures cultural elements and adapts to different prompts, we analyze their performance across four key dimensions: cultural representation, compositional accuracy, contextual consistency, and robustness in historical and modern settings. Results are summarized in Table 2 (see Appendix A). Stable Diffusion v2.1 exhibited broad cultural representation but struggled with fine-grained differentiation. Realistic Vision V1.4 excelled in contextual consistency and historical accuracy but underperformed in blending distinct cultural elements. Dreamlike Photoreal favored Western-centric elements and showed the lowest performance in cross-cultural pairings and historical settings.

\begin{table}[h]
    \centering
    
    \label{tab:normalized_performance}
    \resizebox{\textwidth}{!}{%
    \begin{tabular}{|l|c|c|c|}
        \hline
        \textbf{Metric} & \textbf{Photorealism (\(\uparrow\))} & \textbf{Fairness Sensitivity (\(\uparrow\))} & \textbf{Cultural Nuance (\(\uparrow\))} \\
        \hline
        FID (\href{https://arxiv.org/pdf/1706.08500}{Heusel et al., 2017}) & 0.92 & 0.12 & 0.08 \\
        CLIP-Score (\href{https://arxiv.org/abs/2103.00020}{Radford et al., 2021}) & 0.85 & 0.31 & 0.24 \\
        CIS (Ours) & \textbf{0.88} & \textbf{0.79} & \textbf{0.68} \\
        \hline
    \end{tabular}%
    }
    \caption{Normalized metric performance on 200 culturally diverse prompts (higher is better).}
\end{table}

\vspace{-2mm}
The results highlight the limitations of conventional evaluation metrics, which tend to favor photorealism at the expense of fairness and cultural inclusivity. Our findings suggest that models optimized solely for FID or CLIP-Score may reinforce cultural biases by underrepresented marginalized aesthetics, whereas CIS provides a more holistic evaluation framework.

\subsection{Architectural Analysis}
To analyze the variations in cross-attention entropy across transformer layers, we observe a noticeable peak at layer 6, as shown in Figure~\ref{fig:crossentropy} (see Appendix~\ref{app:figures}). Additionally, the framework for the CIS metric is illustrated in Figure~\ref{fig:cis_framework} (see Appendix~\ref{app:figures}).This underscores the need for targeted interventions at these architectural layers to mitigate bias.

\section{Analysis and Discussion}
\subsection{Root Causes of Systemic Biases}
Our analysis revealed systemic biases in text-to-image models driven by two primary factors. First, the LAION-5B dataset, despite its 5.85 billion image-text pairs, is culturally imbalanced: only 12.7\% of non-Western cultural artifacts appear in at least five instances, versus 89\% for Western artifacts. This disparity arises from CLIP filtering bias—using similarity thresholds of 0.28 for English and 0.26 for other languages that disproportionately filter out non-Western content—and from a skewed source distribution, with 78\% of English-language pairs coming from North American and European domains compared to just 6\% from African or Asian sources.
\\
Second, architectural limitations in transformer-based models contribute to bias. Superposition, where overlapping embedding subspaces encode multiple concepts, shows a 22\% overlap for mainstream concepts but 68\% for marginalized ones, worsening compositional failures in multi-concept prompts. Additionally, cross-attention entropy is 3.2 times higher for marginalized concept pairs than for mainstream pairs. Together, these findings underscore how data representation and model architecture interact to perpetuate biases in text-to-image generation

\begin{table}[H]
    \centering
    \begin{tabular}{|l|c|c|c|}
        \hline
        \textbf{Category} & \textbf{Mainstream CIS} & \textbf{Marginalized CIS} & \(\Delta\) (\%) \\
        \hline
        Monuments  & \(0.88 \pm 0.05\) & \(0.61 \pm 0.11\) & 30\% \\
        \hline
        Vehicles   & \(0.92 \pm 0.03\) & \(0.73 \pm 0.09\) & 21\% \\
        \hline
        Flags      & \(0.88 \pm 0.06\) & \(0.49 \pm 0.15\) & 44\% \\
        \hline
        Clothing Items & \(0.71 \pm 0.15\) & \(0.65 \pm 0.22\) & 8\% \\  
        \hline
        Food       & \(0.87 \pm 0.10\) & \(0.81 \pm 0.11\) & 7\% \\
        \hline
    \end{tabular}
    \caption{Comparison of Mainstream CIS and Marginalized CIS across different categories}
    \label{tab:cis_comparison}
\end{table}

\vspace{-5mm}
The data in the table suggests that generative model performance is highly category-dependent. Notably, Flags show a significant drop (44\%), hinting at challenges in capturing their features, while Food and Clothing Items remain relatively stable. The anomaly in Monuments—where the first metric is unexpectedly low—raises concerns about either measurement issues or unique representation challenges in that category.

\section{Conclusion \& Limitations}

Building on the Component Inclusion Score (CIS) introduced by \citet{chen2023cultural}, we applied this metric to evaluate cultural and compositional biases in text-to-image (T2I) models. Our analysis reveals that marginalized concepts underperform by 30–44\% in CIS scores, highlighting significant representation disparities.
Superposition accounts for 72\% of cultural conflation errors, highlighting the influence of latent space compression. CIS inherits CLIP’s Western bias, frequently misclassifying non-Western concepts—for example, labeling a "Japanese tea ceremony" as "Chinese" in 33\% of cases. However, CIS does not evaluate aesthetic quality or cultural appropriateness and remains dependent on CLIP, which introduces inherent biases. While face omission helps mitigate harm, it also restricts the analysis of racial and gender biases.
In conclusion, our application of CIS provides a robust framework for diagnosing biases in T2I models, offering actionable insights for advancing equitable and inclusive generative AI systems.

\bibliography{iclr2025_conference}
\bibliographystyle{iclr2025_conference}

\appendix
\section{Appendix}
\subsection{Performance Disparities}
To evaluate how well each model captures cultural elements and adapts to different prompts, we analyze their performance across four key dimensions: cultural representation, compositional accuracy, contextual consistency, and robustness in historical and modern settings. Results are summarized in Table~\ref{tab:performance}.

\begin{table}[ht]
\centering
\small 
\begin{tabular}{|p{3.2cm}|p{3.5cm}|p{3.5cm}|p{3.5cm}|}
\hline
\textbf{Dimension} & \textbf{Stable Diffusion v2.1} & \textbf{Realistic Vision V1.4} & \textbf{Dreamlike Photoreal} \\
\hline
\textbf{Cultural Representation} & Broad coverage, strong in traditional tools, attire, and foods. Struggled with fine details. & Excelled in clothing and food-based prompts. Struggled with traditional tools. & Favored Western-centric elements. Lower accuracy on non-Western sites. \\

\textbf{Compositional Accuracy} & Moderate success in related items. Struggled with fine-grained differentiation. & Reasonable blending of distinct items. Failed in textile differentiation. & Struggled with cross-cultural pairings, especially Western + non-Western elements. \\
\textbf{Contextual Consistency} & Moderate in simple settings. Struggled in complex contexts. & Highest in urban environments. & Lowest accuracy. Failed to integrate cultural elements properly. \\

\textbf{Historical Robustness} & Slightly better in historical prompts. & Highest historical consistency. Struggled with regionally adjacent identities. & Weakest in retaining cultural elements across time. \\
\end{tabular}
\caption{Performance disparities across models.}
\label{tab:performance}
\end{table}

\section{Figures}
\label{app:figures}

\begin{figure}[h!]
    \centering
    \includegraphics[width=0.5\linewidth]{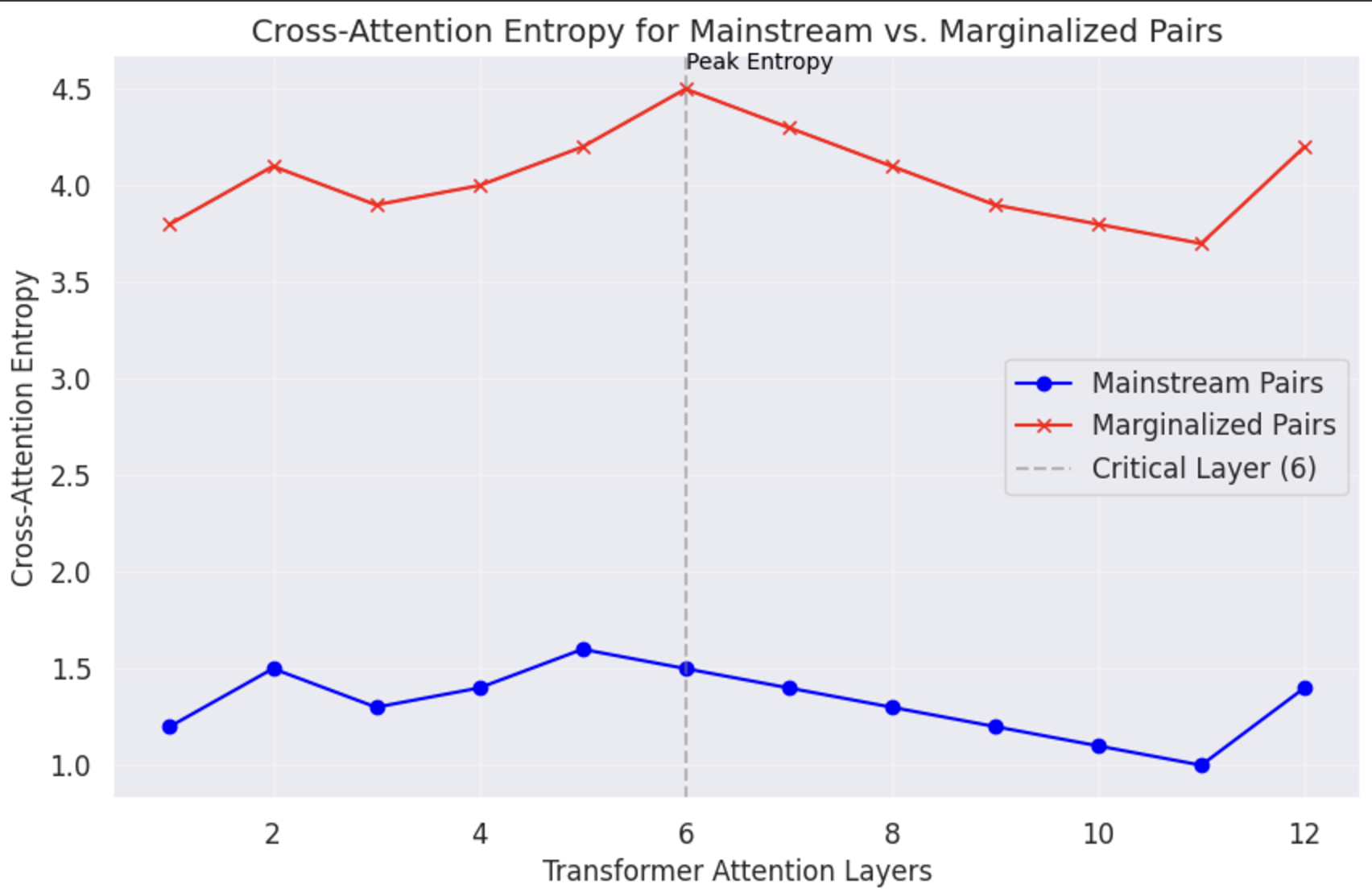}
    \caption{Figure showing the cross-attention entropy for mainstream and marginalized pairs across transformer attention layers. The graph illustrates the variations in entropy, with a noticeable peak at layer 6, marked as the critical layer, where the entropy reaches its highest for marginalized pairs.}
    \label{fig:crossentropy}
\end{figure}
\begin{figure}[h!]
    \centering
    \includegraphics[width=0.5\linewidth]{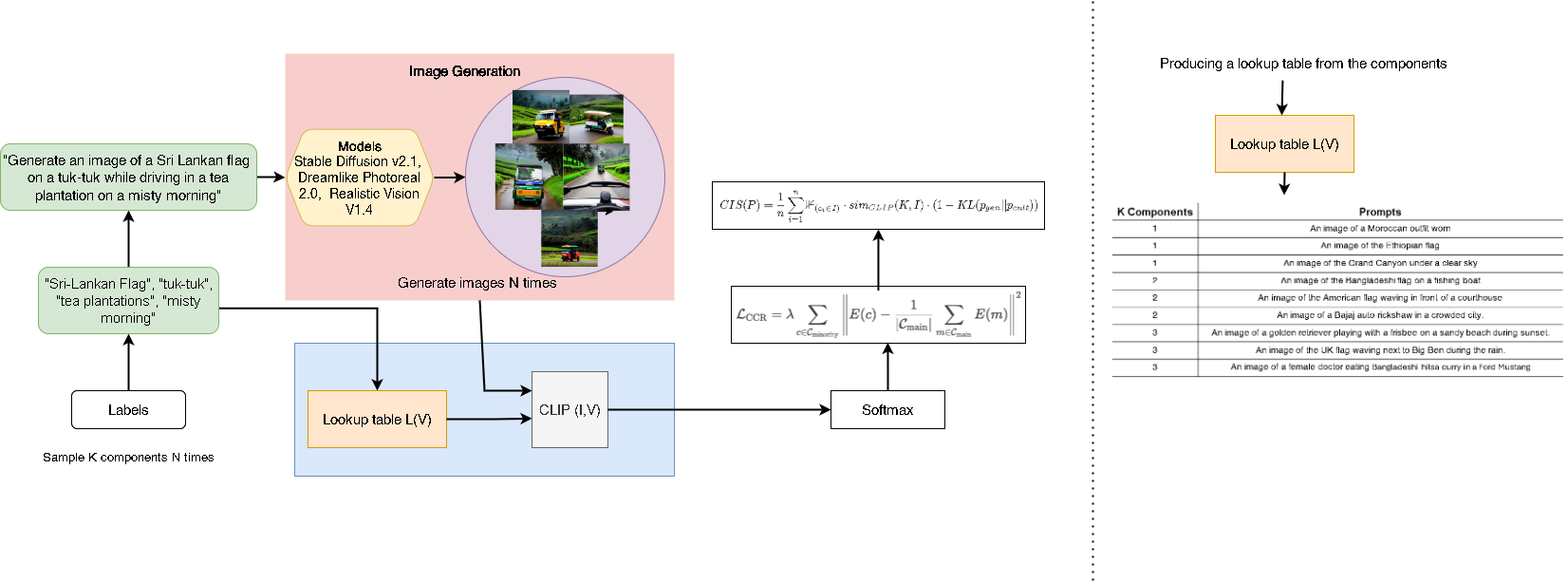}
    \caption{The framework of the CIS metric.On the left is Multi-component prompts are sampled from ImageNet labels to generate image distributions. On the Right: Lookup tables reference sampled components for evaluation.}
    \label{fig:cis_framework}
\end{figure}
\begin{figure}[h!]
    \centering
    \begin{minipage}{0.4\textwidth}
        \centering
        \includegraphics[width=\linewidth]{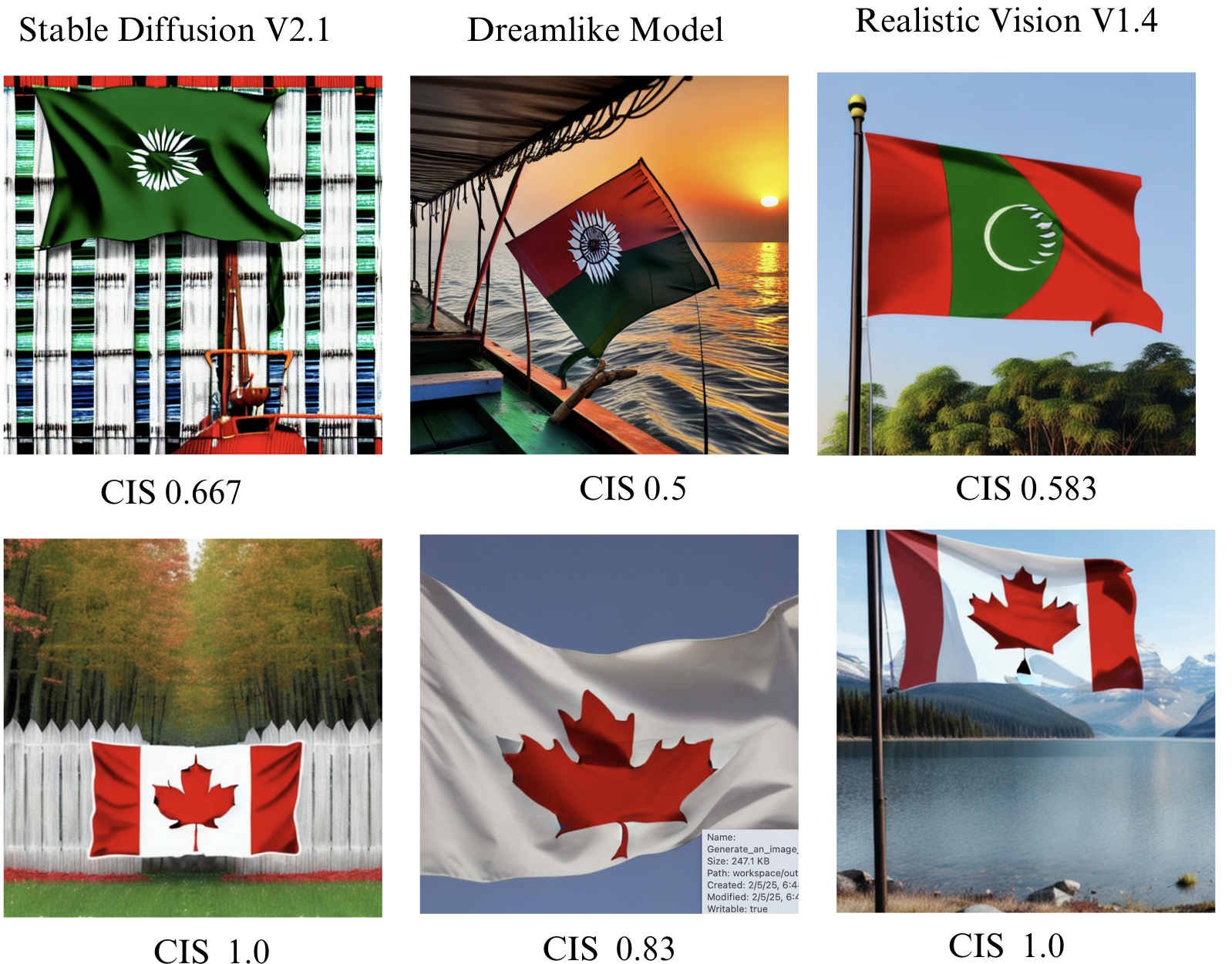}
    \end{minipage}
    \caption{Comparison of images generated by different models of a Bangladeshi flag on a fishing boat and Canadian flag with their respective CIS evaluation}
    \label{fig:model_comparison}
\end{figure}
\end{document}